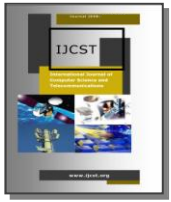

# Recurrent Neural Network Method in Arabic Words Recognition System

Yusuf Perwej

*Abstract*– The recognition of unconstrained handwriting continues to be a difficult task for computers despite active research for several decades. This is because handwritten text offers great challenges such as character and word segmentation, character recognition, variation between handwriting styles, different character size and no font constraints as well as the background clarity. In this paper primarily discussed Online Handwriting Recognition methods for Arabic words which being often used among then across the Middle East and North Africa people. Because of the characteristic of the whole body of the Arabic words, namely connectivity between the characters, thereby the segmentation of An Arabic word is very difficult. We introduced a recurrent neural network to online handwriting Arabic word recognition. The key innovation is a recently produce recurrent neural networks objective function known as connectionist temporal classification. The system consists of an advanced recurrent neural network with an output layer designed for sequence labeling, partially combined with a probabilistic language model. Experimental results show that unconstrained Arabic words achieve recognition rates about 79%, which is significantly higher than the about 70% using a previously developed hidden markov model based recognition system.

*Index Terms*– Recurrent Neural Networks (RNN), Arabic Word, Recognition, Feature Extraction and Language Model

## I. INTRODUCTION

TODAY, character recognition is one of the most challenging tasks and exciting areas of research in computer science. Indeed, despite the growing interest in this field, no satisfactory solution is available [1]. The difficulties encountered are numerous and include the huge variability of handwriting such as writer variabilities writing environment (pen, sheet, support, etc.) The overlap between the characters, and the ambiguity that makes many characters unidentifiable without referring to context. The Arabic script evolved from the Nabataean Aramaic script. It has been used since the 4$^{th}$ century AD, but the earliest document, an inscription in Arabic, Syriac and Greek dates from 512 AD [2]. The Aramaic language has fewer consonants than Arabic, so during the 7th century new Arabic letters were created by adding dots to existing letters in order to avoid ambiguities. The Arabic one of the six official languages of the United Nations, Arabic is a Semitic language spoken by between 300 and 400 million native speakers, and a further 250 million non-native speakers, in nearly twenty countries in the Middle East and North Africa [3]. Arabic is quite a challenging language to learn for a number of reasons: it uses a number of sounds pronounced way back in the throat that can be tricky for speakers of English and other languages [4]. It is written with a cursive alphabet running from right to left in which the letters change shape depending on their position in a word [5].

In this paper we make use of the local features and spatial features for feature selection. Local features come from online information and these features provide information on the dynamics of writing at a very high resolution. While these features are good for representing very local structures in the handwritten strokes, they do not necessarily represent the global structural properties well. So we add the spatial features from offline information. We are proposing a Recurrent Neural Network (RNN) [6] to online [7] handwriting Arabic word recognition. The RNN architecture is bidirectional Long Short- Term Memory [8], chosen for its ability to process data with long time dependencies [9]. The RNN has used the recently introduced connectionist temporal classification output layer [10], which was specifically, designed for labeling unsegmented sequence data. An algorithm is introduced for applying grammatical constraints to the network outputs, thereby providing word level transcriptions. Arabic word is a kind of spelling characters, which has a very special written structure different from English, Hindi and other characters. It is written from right to left from top to bottom, all letters are connected together to form a horizontal backbone [11], which makes the segmentation of Arabic letter very difficult, and every letter may have different shapes in different positions. All these characteristics bring many difficulties to recognition.

## II. RECURRENT NEURAL NETWORKS

Neural Network (NN) mimics biological information processing mechanisms. They are typically designed to perform a nonlinear mapping from a set of inputs to a set of outputs. Neural Network is developed to try to achieve the biological system type of performance using a dense Inter

---

Corresponding author is a research scholar Ph. D (Computer Science) from Department of Computer Science & Engineering, CMJ University , Shilong , Meghalaya , India (Email : yusufperwej@gmail.com)

connection of simple processing elements analogous to biological neurons.

Recurrent neural networks have been an important focus of research and development during the 1990's. They are designed to learn sequential or time varying patterns. A recurrent net is a neural network with feedback (closed loop) connections [12]. Recurrent neural network techniques have been applied to a wide variety of problems. Simple partially recurrent neural networks were introduced in the late 1980's by several researchers including Rumelhart, Hinton, and Williams [13] to learn strings of characters. Many other applications have addressed problems involving dynamical systems with time sequences of events. Learning is a fundamental aspect of neural networks and a major feature that makes the neural approach so attractive for applications that have from the beginning been an elusive goal for artificial intelligence. The added complexity of the dynamical processing in recurrent neural networks from the time-delayed update of the input data requires more complex algorithms for representing the learning. The dynamics of a recurrent neural network can be continuous or discrete in time. However, the simulation of a continuous-time recurrent neural network in digital computational devices requires the adoption of a discrete-time equivalent model.

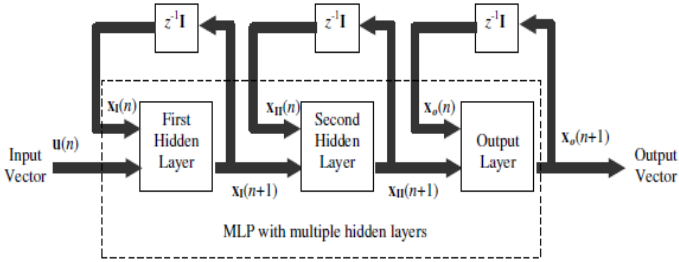

Fig. 1: The structure of recurrent multilayer perceptron (RMLP)

Analysis and synthesis of recurrent neural networks of practical importance are a very demanding task, and second-order information should be considered in the training process. They present a low-cost procedure to obtain exact second order information for a wide range of recurrent neural network architectures. They also present a very efficient and a generic learning algorithm, an improved version of a scaled conjugate gradient algorithm that can effectively be used to explore the available second order information. A schematic 3-layer diagram for RMLP is shown in figure 1, where RMLP can be considered as a feed forward network augmented by recurrent synapses. Generally, a RMLP has one or more recurrent layer, i.e., There exist layer that receives recurrent inputs from themselves and other layers. In this exemplary RMLP, recurrent inputs to a certain layer are solely composed of the action potentials of that layer at the previous time step. Let $x_I(n)$ and $x_{II}(n)$ denote the output of the first and second hidden layer, respectively, and $x_o(n)$ be the output of the output layer and $u(n)$ denotes the input vector. Than the operational principles of the RMLP given in figure 1 can be mathematically expressed by the following coupled equations.

Where $\varphi_I(.,.)$, $\varphi_{II}(.,.)$ And $\varphi_o(.,.)$ are the activation functions of the first hidden layer, second hidden layer, and output layer, respectively; $w_I$, $w_{II}$ and $w_o$ denote the weight matrices of the first hidden layer, second hidden layer, and output layer, respectively.

$$\mathbf{x}_I(n+1) = \varphi_I(\mathbf{w}_I \cdot \begin{bmatrix} \mathbf{x}_I(n) \\ u(n) \end{bmatrix})$$

$$\mathbf{x}_{II}(n+1) = \varphi_{II}(\mathbf{w}_{II} \cdot \begin{bmatrix} \mathbf{x}_{II}(n) \\ \mathbf{x}_I(n) \end{bmatrix})$$

$$\mathbf{x}_o(n+1) = \varphi_o(\mathbf{w}_o \cdot \begin{bmatrix} \mathbf{x}_o(n) \\ \mathbf{x}_{II}(n) \end{bmatrix})$$

## III. CONNECTIONIST TEMPORAL CLASSIFICATION

Connectionism within cognitive science is a theory of information processing. Unlike system which use explicit often logical rule arrange in a hierarchy to manipulate symbol in a serial manner, Connectionist system relies on parallel processing of sub symbol using statistical property instead of logical rules to transform information. Connectionists base their model upon the known neurophysiology of the brain and attempt to incorporate those function properties thought to be required for cognition. The functional properties of the brain that is required for information processing? The Connectionists adopt the view that the basic building block of the brain is the neuron. The neuron has six basic function properties. It is an input device receiving signal from the environment or other neurons. It is an integrative device integrating and manipulating the input. In other words it is a conductive device conducting the integrated information over distance. It is an input device sending information to the other neurons [14]. It is a computational device mapping one type of information into another and it is a representational device subserving the formation of internal representations.

Connectionism are characterized by computationally powerful network that can be fully trained. These networks often hailed as providing a simple universal learning mechanism for cognition. The learning algorithm embodies within Connectionist models have created very powerful information processing they are both universal functions [15] approximators and arbitrary pattern classifiers. Connectionism has a very long past. In fact the origin of Connectionist ideas to the early Greek philosopher, Aristotle and his idea of mental associations. These ideas were elaborated by the British empiricists and then naturally extended by the founders of psychology.

Although it might be argued that these past researchers were not true Connectionists in today's terms the ideas they put forth in the disciplines of philosophy, psychology, Neuropsychology, mathematics and computer science are fully embodied within today Connectionism.

In this paper the output representation that allows a recurrent neural network to be used for CTC. The network can then be used a classifier by selecting the most probable labeling for a given input sequence. Connectionist temporal classification (CTC) is an objective function designed for sequence labeling with recurrent neural network. But the major problem with the traditional objective functions for recurrent neural network is that they require individual targets for each point in the data sequence, which in turn requires the boundaries between segments with different labels to be pre-determined. The Connectionist Temporal Classification output layer [10] solves this problem by allowing the network to choose the location as well as the class of each label.

A CTC output layer contains one more unit than there are elements in the alphabet N of labels for the task. The activations of the first N units are interpreted as the probabilities of observing the corresponding labels at particular times. The activation of the extra unit is the probability of observing a blank in other words no label. The combined output sequence estimates the joint probability of all possible alignments of the input sequence with all possible labeling. The total probability of any one label sequence can then be found by summing the probabilities of its different alignments.

We require an efficient way of calculating the conditional probabilities of individual labeling. This can be solved by a dynamic programming algorithm, similar to the forward-backward algorithm for hidden Markov models [16]. The objective function for CTC is the negative log probability of the network correctly labeling the entire training set. The T is a training set, consisting of pairs of input and target sequences (i, t), where target sequence t is at most as long as input sequence i. The objective function is given below:

$$O^{CTC} = -\sum_{(i,t) \in T} Nm(p(t|i))$$

We have described an output representation that allows recurrent neural networks to be used for $O^{CTC}$. We now derive an objective function for training CTC networks with gradient descent. The weight gradients can be calculated with standard backpropagation through time. Once the network is trained, we would ideally label some unknown input sequence a by selecting the most probable labeling N`.

$$N` = \frac{\arg\max}{N} p(N|i) \quad (1)$$

## IV. LANGUAGE MODEL

A language model is a function, or an algorithm for learning such a function, that captures the salient statistical characteristics of the distribution of sequences of words in a natural language typically allowing one to make probabilistic predictions of the next word given preceding ones. For example, the language model based on a big English newspaper archive is expected to assign a higher probability to "these broking houses provide a" than to "thebr oseking hou sesprose avide", because the words in the former phrase (or word pairs or word triples if so-called N-GRAM MODELS are used) occur more frequently in the data than the words in the latter phrase. For information retrieval, typical usage is to build a language model for each document. At search time the top ranked document is the one which language model assigns the highest probability to the appropriate query.

In the 1990's language models were applied as a general tool for several natural language processing applications, such as part-of-speech tagging, machine translation, and optical character recognition. Language models were applied to information retrieval by a number of research groups in the late 1990's [17]. They became rapidly popular in information retrieval research. We can express these constraints by mending the probabilities in (1) to be conditioned on some probabilistic grammar (gram), as well as the input sequence i.

$$N` = \frac{\arg\max}{N} p(N|i, gram)$$

$$p(N|i, gram) = \frac{p(N|garm) p(N|i) p(i)}{p(N) p(i|gram)}$$

The N contains only dictionary words, can be incorporated by setting the probability of all sequences that fail to meet them in 0. Where we have used the fact that i is conditionally independent of gram given N. If we assume i is independent of gram and all label sequences are equally probable prior to any knowledge about the input or the grammar, we can drop the p(N) term in the divisor to get. The gram consists of a dictionary D containing W words, and a set of $W^2$ bigram (w | ŵ) that define the probability of making a regeneration from word ŵ to word w. The probability of any labeling that does not form a sequence of dictionary words is 0.

$$N` = \frac{\arg\max}{N} p(N|gram) p(N|i)$$

## V. PROPOSED ARABIC WORD RECOGNITION SYSTEM

In this section, the proposed Arabic word recognition system is described. A typical handwriting recognition system consists of Input (Image Acquisition), Preprocessing, feature extraction, classifier and post processing result. The general schematic diagram of the Arabic word recognition system is shown in Figure 2. In input (Image acquisition), the word recognition system acquires a scanned image as an input image. The image should have a specific format such as JPEG, BMP, GIF, PNG etc. This image is acquired through a scanner, digital camera or any other suitable digital input device.

Preprocessing plays an important role in word recognition system as with any other pattern recognition task. The various tasks performed on the image in the preprocessing stage are shown in Fig. 3. This stage includes noise removing, averaging binarization process. Handwritten characters show various undesirable effects like small unwanted strokes, gaps or breaks which occur due to binarization. Many a times when a character is handwritten, it exhibits lesser width at the curvature than in other parts of the character. This point is more likely to break during binarization.

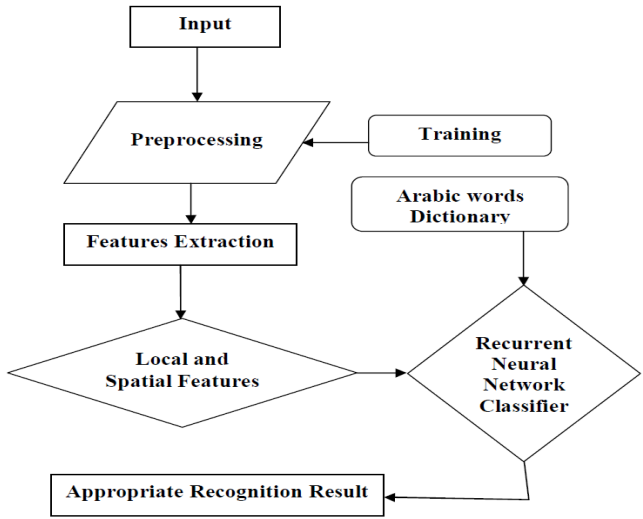

Fig. 2: Arabic word recognition system

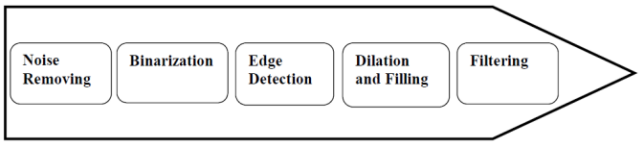

Fig. 3: The Preprocessing of Arabic word recognition of image

A 3×3 averaging operator is implemented before binarization, which blurs the image resulting into bridging small gaps and retaining the actual shape of the character. It also removes pepper noise. The unwanted strokes occur most often between the pen lifting and placing points and their occurrence depend upon the writing style and the ink viscosity. These strokes may result into unwanted feature detection after binarization. In order to avoid this, the binarized image should be cleaned. This is done by using morphological technique. Morphological technique removes thin protrusions, breaks thin connections and smoothes the object contour. For example, $p^m = p^m + 1 = \ldots = p^m + n$ for some m and n. When computing differential features, these co-occurrences can reason disturbing uniqueness. We are erasing the extra occurrences $p^m$, $p^m + 1 \ldots p^m + n$. After that the rustication of repeated samples, the input signals must be filtered in order to reduce noise. In this paper, we have applied a Gaussian filter freely to each of the x and y coordinates of the point sequence.

$$X_l^{filt} = \sum_{m=-3\delta}^{3\delta} w_m x_{l+m}^{real}$$

$$w_m = \frac{a - \frac{m^2}{2\delta^2}}{\sum_{j=-3\delta}^{3\delta} a^{-\frac{j^2}{2\delta^2}}}$$

In this paper we are using a Slant correction process for Arabic word recognition because the Arabic word has a horizontal baseline and write right to left. In the preprocessing process is erased the repeated Input sample and filtering in other words smoothing the input sample after that Slant correction & detection and size normalization. After preprocessing the input sample was moved into the feature extraction module.

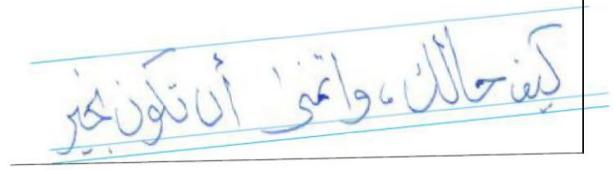

Fig. 4: Arabic words slant correction

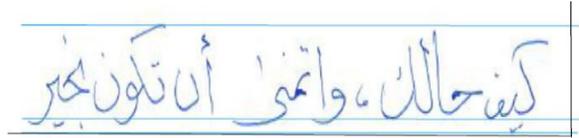

Fig. 5: Arabic words slant correction after transform

The feature extraction has been highly inspired by the human reading process that considers the global high level words shape [18]. Feature extraction stage in character recognition as in any pattern recognition task plays a major role in ameliorate the recognition precision [19]. Because the features are extracted from binary characters. Thus, the characteristics used for classification lie solely in the shape variations. Many characters are similar in shape or slight variations in the writing style may result into misclassification [20]. The features selected should tackle all these problems. Moreover, a single feature extraction and classification stage may recognize a character which may not be recognized by the other feature extractor and classifier combination. Hence a hybrid system is needed that can recognize the characters over a wide range of varying conditions [21]. The cropped characters in each structural class are resized to a fixed size before extracting the features. The collection of extracting features can be divided into two categories. The first category consists of features extracted for each point pi considering the vicinal of pi with respect to time. Another category takes the off-line matrix representation of the handwriting into the computation. In this paper we discovered the best character recognition rate in combination with our categorization using the sequence [ $X_m, Y_m, \theta_m$ ] as the first class features. $X_m$ and $Y_m$ is coordinate points, $\theta_m$ is the tangent slope angle at point m. Where $\theta_m = N_m$ and J= -1 and arge the phase of the complex number below.

$$N_m = \arg e((X_{m+1} - X_{m-1}) + j*(Y_{m+1} - Y_{m-1}))$$

The features of the second category are all computed using a two-dimensional matrix symbolizing the off-line edition of the data. The number of points up and down the corpus in the X neighborhood of pi and the two-dimensional neighborhood of pi transforms into a 3 × 3 map these resulting nine values are taken as features. The proposed system is implemented using MATLAB on Intel Core 2 Duo CPU, running at 2.40 GHz with 2 GB RAM.

## VI. EXPERIMENTAL RESULTS

We are used handwriting Arabic words database, which contains 20000 word instances from a 7500 Arabic a word dictionary and the training set consists of 15000 Arabic words including all different types of words in the database. We are randomly selected 2000 Arabic words for testing sample from remained samples. The connectionist temporal classification network used the recurrent neural network architecture. The forward and backward hidden layers each contained 50 single cell memory blocks. The all input layers were fully connected to the hidden layers, which were fully connected to themselves and the output layer. The output layer contained 42 Arabic word characters plus the 1 blank label and the preprocessed data, there were 7 inputs. In this paper we are using for the logistic sigmoid and cell activation functions in the range of [0, 1] was used for the gates. The both input representations, the data was normalized so that each input standard deviation 1 and had mean 0 of the training set. Training was carried out with back propagation through time and online gradient descent in other words weight updates dynamically after every training, using a learning rate of $5-2$ and a speed of 0.8. The weights were initialized with a flat inconsequent distribution in the range of $0.1$ and $-0.1$.

Gaussian noise was added to the inputs with a standard deviation of 0.55 to be rectified generalized during training time. Before starting training network activations were reset to 0 of each example and training was stopped until performance achieved the validation set. In this paper results are partitions into two groups, one is writing Arabic words in coerce type. We are compared connectionist temporal classification with hidden Markov model system and record the Arabic word recognition rate with a language model and a dictionary.

Table 1 indicates the comparison of the connectionist temporal classification and hidden Markov model system, along with the unconstrained experimental results obtained with the proposed system. The recognition accuracy of 79% is a significant improvement. If we comply with Arabic words writing stroke order, the recognition rate can achieve 88% (Table 2). According to the experimental results, as far as we know, RNN classifier for handwriting Arabic word recognition is feasible and outperforms any other single model.

## VII. CONCLUSION

Finally, the word recognition has been one of the active and challenging research areas in the field of image processing and pattern recognition. In the beginning we discuss the technology of recurrent neural network based on online and offline information. In this paper we are using method fits naturally into the existing framework of neural network classifiers, and are derived from the same probabilistic principles. It obviates the need for pre-segmented data, and allows the network to be trained directly for sequence labeling. Experimental results show that the recurrent neural network is distinctly superior to the other hidden Markov model in recognizing the handwritten Arabic words. The proposed system will find useful applications in words recognizing the handwritten names, reading documents and conversion of any handwritten document in structural text form etc. Moreover, without requiring any task specific knowledge, it has outperformed a hidden Markov models on a real life temporal classification problem. In the forthcoming, we will continue being rectified our system for higher recognition rate of unconstrained cursive and writer independent. Therefore, the combination of a feature extraction technique with recurrent neural network may provide better solutions for Arabic word recognition.

Table 1: The unconstrained Arabic word recognition rate

| Using Method | Using Language Model | Arabic Word Recognition Rate Top - 1 | Arabic Word Recognition Rate Top - 5 | Arabic Word Recognition Rate Top – 10 |
|---|---|---|---|---|
| Hidden Markov Model (HMM) | Yes | 68% | 74% | 83% |
| Connectionist Temporal Classification(CTC) | Yes | 79% | 88% | 93% |
| Connectionist Temporal Classification(CTC) | No | 71% | 78% | 86% |

Table 2: The unconstrained Arabic word recognition rate (It contains complied with stroke order and no delayed stroke)

| Using Method | Using Language Model | Arabic Word Recognition Rate Top - 1 | Arabic Word Recognition Rate Top - 5 | Arabic Word Recognition Rate Top – 10 |
|---|---|---|---|---|
| Hidden Markov Model (HMM) | Yes | 71% | 83% | 89% |
| Connectionist Temporal Classification(CTC) | Yes | 88% | 91% | 96% |
| Connectionist Temporal Classification(CTC) | No | 73% | 84% | 88% |

## REFERENCES

[1] R. Plamondon, S. N. Srihari, "On-line and off-line handwriting recognition: A comprehensive survey", IEEE Trans. Pattern Anal. Mach. Intell., vol. 22(1), pp 63–84, 2000.

[2] M. Cheriet, M. Beldjehem. "Visual Processing of Arabic Handwriting:Challenges and New Directions", Summit on Arabic and Chinese handwriting (SACH'06), Washington-DC, USA, pp 129-136, 2006.

[3] J. Dichy ,"On lemmatization in Arabic. A formal definition of the Arabic entries of multilingual lexical databases" , ACL

39th Annual Meeting. Workshop on Arabic Language Processing; Status and Prospect. Toulouse, pp 23-30, 2001.
[4]  I. A. Al-Sughaiyer and I. A. Al-Kharashi, "Arabic Morphological Analysis Techniques," J. of the Amer. Soc. For Inf. Sc. And Tech., vol 55, No. 3, pp 189-213, 2004.
[5]  MA. Attia , T. Salakoski , F. Ginter , S. T. Pyysalo ,"Accommodating Multiword Expressions in an Arabic LFG Grammar " , In Finland Springer-Verlag Berlin Heidelberg, vol 4139, pp 87 – 98, 2006
[6]  M. Schuster and K. K. Paliwal, "Bidirectional recurrent neural networks", IEEE Transactions on Signal Processing,vol 45, pp 2673–2681, November 1997.
[7]  A. M. Schaefer, S. Udluft, and H. G. Zimmermann, "Learning long-term dependencies with recurrent neural networks," Neurocomputing, vol. 71, No. 13-15, pp 2481–2488, 2008.
[8]  A. Graves , J. Schmidhuber, "Framewise phoneme classification with bidirectional LSTM and other neural network architectures," Neural Networks, vol. 18, No. 5-6, pp 602–610, 2005.
[9]  M. W¨ollmer, B. Schuller, F. Eyben, and G. Rigoll, "Combining long short-term memory and dynamic bayesian networks for incremental emotion-sensitive artificial listening", IEEE Journal of Selected Topics in Signal Processing, Special Issue on Speech Processing for Natural Interaction with Intelligent Environments, vol 4(5), pp 867–881, 2010.
[10] A. Graves, S. Fernandez, F. Gomez, and J. Schmidhuber, "Connectionist temporal classification: Labelling unsegmented data with recurrent neural networks", In Proc. of 23rd Int. Conf. on Machine Learning, Pittsburgh, USA, 2006.
[11] D. Kamir , N. Soreq , Y. Neeman , " A Comprehensive NLP System for Modern Standard Arabic and Modern Hebrew" , Proceedings of the workshop on Computational Approaches to Semitic Languages in the 40th Annual Meeting of the Association for Computational Linguistics (ACL- 02). Philadelphia, PA, USA., 2002.
[12] L. Fausett, "Fundamentals of Neural Networks" , Prentice Hall,  Englewood Cliffs, New Jersey, 1994.
[13] D. E. Rumelhart, G. E. Hinton, and R. J. Williams, " Learning internal representations by error propagation, in Parallel Distributed Processing: Explorations in the Microstructure of Cognition", MIT Press, Cambridge, vol. 1, pp 45–76, 1986.
[14] Y. Dudai, "The Neurobiology of Memory" , Oxford University Press, Oxford , 1989.
[15] R.P. Lippman, "An introduction of computing with neural nets, IEEE ASSP Magazine, vol.4, pp 4-22, April 1987.
[16] L. R. Rabiner, "A tutorial on hidden markov models and selected applications in speech recognition", Proc. IEEE, vol. 77, No. 2,  pp 257—286, FEBRUARY 1989.
[17] J. M. Ponte , W. Bruce Croft., "A language modeling approach to information retrieval" ,In Proceedings of the 21st ACM Conference on Research and Development in Information Retrieval (SIGIR'98), pp 275–281, 1998.
[18] S. Madhvanath, V. Govindaraju ,"The Role of Holistic Paradigms in Handwritten word Recognition" , IEEE Trans. On Pattern Analysis and Machine Intelligence, vol. 23, No. 2, pp 149-164, February 2001.
[19] R. G. Casey, and E. Lecolinet, "A Survey of Methods and Strategies in Character Segmentation", IEEE Transactions on Pattern Analysis and Machine Intelligence, vol. 18, No. 7, pp 690-706, 1996.
[20] O. D. Trier, A. K. Jain and T. Taxt, "Feature Extraction Methods for Character Recognition A Survey", Pattern Recognition, vol. 29, Issue 4, pp 641–662, April 1996.
[21] S.V. Rajashekararadhya , P. VanajaRanjan, "Efficient zone based feature extraction algorithm for handwritten numeral recognition of four popular south-Indian scripts", Journal of Theoretical and Applied Information Technology, vol.4, No.12, pp 1171-1181, 2008.